%% file: aspdac_sunbin_bass.tex
\documentclass[conference]{IEEEtran}
\IEEEoverridecommandlockouts
\input{setting-ieee}

\graphicspath{{./pic/}{../}}

\usepackage{textcomp}
\usepackage{xcolor}
\usepackage{comment}
\usepackage{booktabs}
\usepackage{multirow}
\usepackage{array}
\usepackage[labelformat=simple]{subcaption}

\usepackage{tabularx}

\usepackage[a4paper, total={184mm,239mm}]{geometry}
\def\BibTeX{{\rm B\kern-.05em{\sc i\kern-.025em b}\kern-.08em
    T\kern-.1667em\lower.7ex\hbox{E}\kern-.125emX}}
    
\definecolor{darkgreen}{rgb}{0.0, 0.5, 0.0} 

\usepackage{multirow}
\usepackage{xcntperchap}
\usepackage{tikz}

\newcommand*\blackcircled[1]{%
    \tikz[baseline=(char.base)]{
        \node[shape=circle,
              draw,
              inner sep=0.2pt,
              minimum size=6pt,
              text=white,
              fill=black] (char) {#1};
    }
}

\begin{document}

\title{\huge
    ParaGate: Parasitic-Driven Domain Adaptation Transfer Learning for Netlist Performance Prediction
    }

\author{
\IEEEauthorblockN{Bin Sun\textsuperscript{1,5}, Jingyi Zhou\textsuperscript{2}, Jianan Mu\IEEEauthorrefmark{1}\textsuperscript{1}, Zhiteng Chao\textsuperscript{1}, 
Tianmeng Yang\textsuperscript{3}, Ziyue Xu\textsuperscript{4}, Jing Ye\textsuperscript{1}, Huawei Li\IEEEauthorrefmark{1}\textsuperscript{1}}
\IEEEauthorblockA{\textsuperscript{1}\textit{State Key Lab of Processors, Institute of Computing Technology, Chinese Academy of Sciences}\\
\textsuperscript{2}\textit{Department of Electronic Engineering, Tsinghua University}\\
\textsuperscript{3}\textit{Peking University}\\
\textsuperscript{4}\textit{School of Information Engineering, Minzu University of China}\\
\textsuperscript{5}\textit{University of Chinese Academy of Sciences}\\
sunbin23@mails.ucas.ac.cn, zhou-jy22@mails.tsinghua.edu.cn, \\
\{mujianan, chaozhiteng20g, yejing, lihuawei\}@ict.ac.cn, \\
youngtimmy@pku.edu.cn, 22012820@muc.edu.cn}
}

\maketitle
\input{doc/abstract}

\input{doc/intro}
\input{doc/prelim}
\input{doc/algo}

\input{doc/result}
\input{doc/conclu}

\newpage
\bibliographystyle{IEEEtran}
\bibliography{ref}

\end{document}

%% file: setting-ieee.tex
\usepackage{blkarray}                                      
\usepackage{algpseudocode}                                 
\usepackage{algorithm}
\usepackage{graphicx}                                      
\usepackage{amsmath}
\usepackage{amssymb}
\usepackage{amsfonts}
\usepackage{amsthm}
\usepackage[mathcal]{eucal}
\usepackage{mathrsfs}
\usepackage{booktabs}
\usepackage{enumerate}
\usepackage{multirow}

\usepackage{color}
\usepackage{cite}                                          
\usepackage{comment}                                       
\usepackage{soul}                                          
\soulregister\cite7
\soulregister\ref7
\soulregister\pageref7
\usepackage{etoolbox}                                      
\usepackage{url}
\usepackage{nth}                                           
\usepackage{bm}                                            
\usepackage{courier}
\usepackage{balance}
\usepackage{threeparttable}
\usepackage{xcolor,colortbl}
\usepackage{footnote}
\usepackage{listings}
\usepackage{setspace}                                      
\usepackage[inline]{enumitem}

\usepackage{verbatim}
\usepackage[bookmarks=false]{hyperref}
\hypersetup{
    colorlinks = true,
    citecolor  = blue,
    linkcolor  = blue,
    urlcolor   = blue,
}
\usepackage{tikz}
\usetikzlibrary{patterns,snakes}
\usetikzlibrary{positioning,calc,fit,decorations.pathmorphing,shapes.geometric, shapes.gates.logic.US, calc}
\usetikzlibrary{arrows,arrows.meta,decorations.markings,shapes,shapes.arrows}
\usetikzlibrary{decorations,decorations.pathreplacing}
\usetikzlibrary{backgrounds}
\usepackage{filecontents}                                  
\usepackage{pgfplots}
\usepackage{pgfplotstable}
\usepackage{scalefnt}
\pgfplotsset{compat=newest}
\usepackage{caption}
\usepackage{pifont}                                        
\usepackage{cleveref}
\Crefformat{figure}{Fig.~#2#1#3}                           
\Crefname{subfigure}{Fig.}{Figs.}
\Crefname{figure}{Fig.}{Figs.}
\Crefformat{table}{TABLE~#2#1#3}                           
\usepackage[figuresright]{rotating}

\definecolor{CUHKorange}{RGB}{244,106,18} 
\definecolor{CUHKblue}{RGB}{0,111,190}    
\definecolor{CUHKgreen}{RGB}{0,127,128}   
\definecolor{CUHKred}{RGB}{228,46,36}     
\definecolor{CUHKyellow}{RGB}{198,148,34} 
\definecolor{CUHKdark}{RGB}{114,44,114}   
\definecolor{CUHKmiddle}{RGB}{144,44,144} 
\definecolor{CUHKlight}{RGB}{167,44,167} 
\definecolor{CUHKpurple}{RGB}{117,15,109}
\definecolor{CUHKgold}{RGB}{221,163,0}
\definecolor{CUHKribbon}{RGB}{244,223,176}
\definecolor{CUHKblack}{RGB}{34,24,21}

\renewcommand{\bf}[1]{\textbf{#1}}
\renewcommand{\tt}[1]{\texttt{#1}}
\renewcommand{\vec}[1]{\boldsymbol{#1}}    

\newcommand{\minisection}[1]{\vspace{.06in}\noindent{\textbf{#1}}.}

\usepackage{tcolorbox}
\tcbuselibrary{skins,breakable}
    {\endtcolorbox}
    {\endtcolorbox}

\crefname{mytheorem}{Theorem}{Theorems}
\crefname{mylemma}{Lemma}{Lemmas}
\crefname{myclaim}{Claim}{Claims}
\crefname{myproperty}{Property}{Properties}
\crefname{mycorollary}{Corollary}{Corollaries}

\algrenewcommand\textproc{\texttt}

\makeatletter
\let\OldStatex\Statex
\renewcommand{\Statex}[1][3]{%
  \setlength\@tempdima{\algorithmicindent}%
  \OldStatex\hskip\dimexpr#1\@tempdima\relax
}
\makeatother

\RequirePackage[normalem]{ulem} 
\RequirePackage{color}\definecolor{RED}{rgb}{1,0,0}\definecolor{BLUE}{rgb}{0,0,1} 

%% file: doc/abstract.tex
\begin{abstract}

In traditional EDA flows, layout-level performance metrics are only obtainable after placement and routing, hindering global optimization at earlier stages. Although some neural-network-based solutions predict layout-level performance directly from netlists, they often face generalization challenges due to the black-box heuristics of commercial placement-and-routing tools, which create disparate data across designs.
To this end, we propose ParaGate, a three-step cross-stage prediction framework that infers layout-level timing and power from netlists. 
First, we propose a two-phase transfer-learning approach to predict parasitic parameters, pre-training on mid-scale circuits and fine-tuning on larger ones to capture extreme conditions. Next, we rely on EDA tools for timing analysis, offloading the long-path numerical reasoning. Finally, ParaGate performs global calibration using subgraph features. Experiments show that ParaGate achieves strong generalization with minimal fine-tuning data: on openE906, its arrival-time 
$R^2$ from 0.119 to 0.897. These results demonstrate that ParaGate could provide guidance for global optimization in the synthesis and placement stages.

\end{abstract}

%% file: doc/intro.tex
\section{Introduction}

As circuit designs scale up, the stage-by-stage approach in traditional EDA flows often leads to slow pipeline-based feedback for final performance metrics, since obtaining timing, power, and other indicators requires completing the entire synthesis and physical design cycle, thus hindering global optimizations \cite{chen2024dawn,fang2025survey,huang2021machine,chang2025large}. Therefore, AI-assisted cross-stage performance prediction has emerged, aiming to provide earlier and more effective guidance. Some recent studies have trained graph neural networks (GNNs) on RTL\cite{fang2023masterrtl,wang2025bridging} or netlists \cite{he2022accurate, guo2022timing, du2024powpredict, zhong2024preroutgnn, wang2023restructure, lyu2024explainable} to model circuit structures and signal propagation, thereby predicting post-layout metrics such as timing and power with strong accuracy on standard benchmarks.

\begin{figure}[tb!]
    \centering
    \subfloat[]{\includegraphics[width=0.83\linewidth]{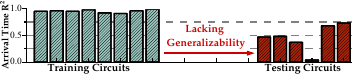} \label{fig:problem_a}}
    \\
    \subfloat[]{\includegraphics[width=0.83\linewidth]{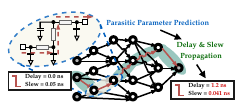} \label{fig:problem_b}}
    \label{fig:problem}
    \caption{(a) End-to-end method lacking generalizability; (b) Potential task coupling in PPA prediction}
\end{figure}

However, one of the limitations of existing solutions lies in their generalizability \cite{chai2023circuitnet,guo2022timing}. As shown in \Cref{fig:problem_a}, we replicated a state-of-the-art GNN-based end-to-end predictor and found that when the test circuit differs significantly from the circuits in the training set (e.g., a larger-scale circuit), its predictive performance worsens.
Some prior studies have achieved favorable performance within known designs by predicting remaining parts based on partial knowledge from the same circuit, or by learning from similar circuits \cite{shen2024deep, davis2021fast}.
However, there are scenarios where the user must predict the PPA of an entirely new circuit. A straightforward approach is to build a training set encompassing numerous diverse circuits; yet, the topological design space of functional units is practically infinite, making it impossible to exhaustively represent all structures.
Therefore, beyond simply expanding the training set, it is also necessary to optimize the prediction methodology.

Predicting layout-level PPA from the netlist is fundamentally physical numerical reasoning based on connectivity relationships. As shown in \Cref{fig:problem_b}, it in fact encompasses two tasks:
Task (1): Predict the parasitic-parameter topology.
For this task, parasitic parameters are primarily determined by the wires between two pins, so the model mainly needs to learn the correspondence between circuit structures and parameters within a smaller local region.
Task (2): Numerical reasoning for arrival time and power based on the topological graph to emulate delay and slew propagation.
However, neural networks are not well-suited for numerical reasoning along long topological paths \cite{zhang2019can, zhang2020efficient}. Moreover, the path topologies of different circuits often differ significantly, further increasing the difficulty of generalization.
In summary, for cross-stage prediction, the significant structural differences in long paths among various circuits—together with the unfriendliness of path-topology-based numerical reasoning to neural networks—make it challenging to train a generalized model.

Based on this analysis, a concise and seemingly promising framework appears to emerge: decoupling end-to-end cross-stage PPA prediction into two steps. The first step is to use a neural network to predict the physical parameters between pins-essentially providing a cross-stage ``forecast'' for the layout level. The second step is to leverage the mature timing reasoning engine in EDA tools to obtain the timing and power metrics.
However, although this decoupling seems promising, achieving cross-stage parasitic parameter prediction for general circuits is still far from trivial. Some studies predict parasitic parameters at the netlist level, but they mainly focus on specific circuit structures, such as SRAM\cite{shen2024deep}. Developing a generalized model capable of predicting across different circuits is more challenging. First, in large-scale digital circuits, the range of parasitic parameters between nodes can span four to five orders of magnitude, leading to data imbalance. 
Second, commercial placement-and-routing tools operate as black boxes using heuristic algorithms for optimization, which work uniquely across different circuit structures and paths. These issues pose challenges for the generalization of prediction models. 
In addition, when modeling the routing, one cannot solely capture the cell information at both endpoints of the wires; it is necessary to incorporate subgraph-level structure to perceive vital information like congestion.

To this end, we propose ParaGate, a cross-stage prediction framework. It takes the netlist as input and predicts the layout-level timing and power metrics.
ParaGate is a three-step predictive framework. At its core is step 1, a two-phase transfer learning framework dedicated to parasitic parameter prediction. We first pre-train on small- and medium-scale circuits to learn general placement-and-routing rules; then we fine-tune the model on data from large-scale circuits with more complex conditions, sampling to capture extreme cases with unique characteristics. In step 2, we employ EDA tools to perform timing analysis. Finally, in step 3, we use a model that incorporates global information to calibrate the results.
Our contributions are summarized as follows:

\begin{itemize}

    \item We propose a three-step framework: using an AI model to predict physical parameters, invoking EDA tools to infer metrics on long paths, and finally performing error correction by incorporating subgraph information.
    \item For physical parasitic parameter prediction, we innovatively propose a “pre-training + fine-tuning” training framework to better capture general patterns and extreme-value distributions.
    \item We also employ a calibration model that captures subgraph features to further refine the prediction metrics.
    \item Experimental results confirm that ParaGate achieves outstanding generalization on unseen samples with only 20\% target domain data for fine-tuning. This is evident in the \textit{openE906} arrival time $R^2$ improving significantly from 0.119 to 0.897, and a total power relative error of just 0.909\% on \textit{BoomTile\_Tiny}.
 
\end{itemize}

%% file: doc/prelim.tex
\section{Preliminary}
\label{sec:prelim}

\subsection{PPA Analysis and Cross-Stage Prediction Models}
Current research on performance, power, and area (PPA) analysis and prediction at the gate-level netlist can be broadly categorized into two types:
The first category focuses on accelerating time-consuming analysis tasks within pre-layout netlists by leveraging machine learning methods. 
The second category primarily concentrates on cross-stage prediction. Among these, representative works include Timer-inspired GNN \cite{guo2022timing}, which draws inspiration from timing analysis tools to realize an end-to-end pre-routing slack prediction framework. LaRC-Timer \cite{he2022accurate} performs delay prediction by rapidly estimating an RC Network and subsequently applying a tree model. Wang $et\ al.$ \cite{wang2023restructure} addresses the impact of topological changes in the netlist caused by back-end restructuring on delay prediction;  Lyu $et\ al.$ \cite{lyu2024explainable} proposed an interpretable and layout-aware end-to-end timing prediction method; PreRoutGNN \cite{zhong2024preroutgnn} proposes a two-step pre-routing timing prediction scheme based on graph autoencoders; and Du $et\ al.$ introduced PowPrediCT\cite{du2024powpredict}, the first cross-stage power prediction model with generalization capabilities. It is noteworthy that current efforts in cross-stage prediction are predominantly based on post-layout information, primarily tackling pre-routing timing and power prediction.

Moreover, in addressing physical information discrepancies, some researchers have pursued alternative avenues: they attempt to directly predict parasitic parameters or other factors contributing to pre/post-layout differences. For example, ParaGraph \cite{ren2020paragraph} uses GNNs to predict parasitic parameters from mixed-signal schematics; GNNs are utilized \cite{shen2024deep} to predict capacitance in SRAM-type circuits; and Mlparest \cite{shook2020mlparest} employs a random forest model \cite{biau2016random} for interconnect resistance and capacitance prediction on analog circuits. However, due to the complexity of parasitic effects, these works are often confined to restricted circuit scales and types. The field still lacks effective parasitic parameter prediction for large-scale general-purpose circuits and subsequent PPA inference.

\begin{figure*}[tb!]
    \centering
    \includegraphics[width=0.95\linewidth]{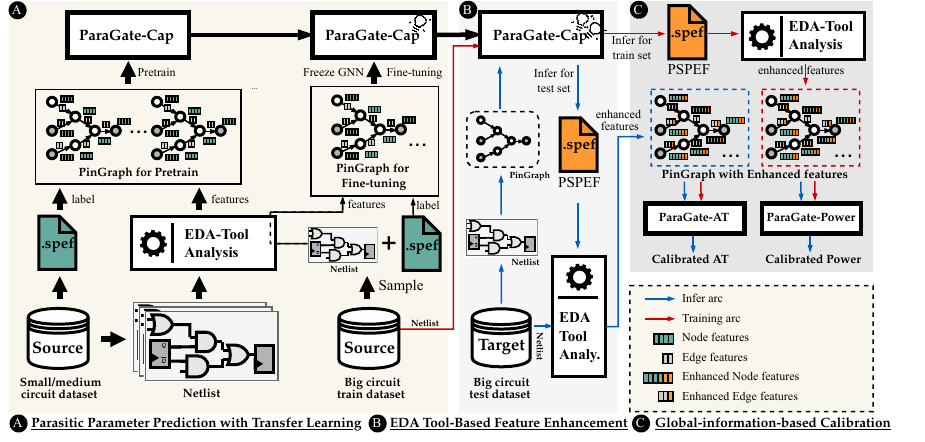}
    \caption{ParaGate framework.}
    \label{fig:framework}
\end{figure*}

\subsection{Graph Neural Network}
Graph neural networks (GNNs) are models designed for graph-structured data. They learn representations by passing and aggregating information between nodes \cite{liu2022introduction,khemani2024review,wu2020comprehensive}, with the message passing mechanism as their core \cite{zheng2024revisiting,he2023message}.

 \minisection{Synchronous and Asynchronous Message Passing}
Synchronous message passing GNNs (SMP-GNNs) update all node representations simultaneously, constraining each node's receptive field to the number of GNN layers. Conversely, asynchronous message passing GNNs (AMP-GNNs) iteratively update node subsets, allowing receptive fields to extend beyond layer counts and capture information from arbitrarily distant nodes until convergence\cite{faber2022asynchronous}.


%% file: doc/algo.tex
\section{ParaGate}
\label{sec:algo}



To enable a generalizable cross-stage timing and power prediction model, we propose ParaGate, a three-step coupled framework illustrated in \Cref{fig:framework}. As shown in \Cref{fig:framework}, the proposed method comprises three steps in the training phase. In step 1, we incorporate circuit structural information to train our parasitic-parameter model, ParaGate-Cap, which involves two phases: model pretraining and subsequent fine-tuning. Next, in step 2, the EDA tool’s inference engine refines PPA metrics at the circuit level. Finally, in step 3, we leverage large-scale circuit training data along with the refined PPA-annotated netlist to adjust timing and power predictions using subgraph-level information.
During inference, for a given input circuit, we first employ the fine-tuned ParaGate-Cap model to predict parasitic parameters. We then feed these predictions into the EDA tool, which propagates timing (arrival time, AT) and power estimates through the circuit topology. Finally, ParaGate-AT and ParaGate-Power are utilized to correct and refine these predictions.
Key innovations of the proposed framework include:

\begin{itemize}
    \item \textbf{Three-step decoupling for cross-stage PPA prediction. (\Cref{fig:framework} \blackcircled{A}- \blackcircled{B}- \blackcircled{C}):} By separating the workflow into three steps, each subproblem can be effectively addressed to improve model generalizability.
    \item \textbf{Parasitic parameter prediction with Transfer Learning (\Cref{fig:framework} \blackcircled{A}):} Our parasitic-modeling approach focuses on local interconnects and inherently learns the behavior of the routing tool. Since routing characteristics can vary significantly across different circuit specifications, our “pretraining + fine-tuning” strategy first learns universal patterns before adapting to extreme cases.
    \item \textbf{Global-information-based Calibration (\Cref{fig:framework} \blackcircled{C}):} Practical routing entails considerations such as regional density, so global circuit context is essential for accurate parameter predictions. To this end, after the initial EDA inference, we propose a subgraph-based model to correct arrival time and power results, further improving prediction precision.
\end{itemize}


\subsection{Upstream - Parasitic Parameter Prediction with Transfer Learning}\label{3.1}

To more accurately represent the netlist, we construct a Pin-Graph to capture the circuit’s structural information, and then employ an SMP-GNN for model fitting. To better learn routing behaviors, we devise a pretraining + fine-tuning framework: the model first acquires generalizable post-routing patterns, and then, via fine-tuning, adapts to task-specific nuances—thereby achieving more precise modeling outcomes.

\begin{figure}[tb!]
    \centering
    \includegraphics[width=0.95\linewidth]{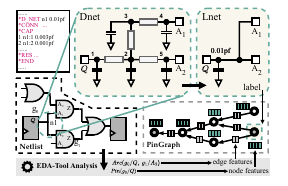}
    \caption{Generation of PinGraph features and label.}
    \label{fig:Cap_predict}
\end{figure}

\minisection{Graph Data Construction}
To accurately capture AT and power propagation, it is critical to traverse the signal pathways within the circuit, which may have multi-fanout structures. Consequently, instead of using a cell-level graph, we adopt a PinGraph. 
Additionally, for the parasitic parameter model, since the netlist before layout lacks physical distances information, we opted for the lumped-capacitance modeling approach among the classic models (distributed model, the lumped-capacitance model, and a combined lumped capacitance-resistance model).
Finally, as depicted in \Cref{fig:Cap_predict}, we build a PinGraph based on the lumped-capacitance framework.
By comprehensively embedding these logical and electrical characteristics of the gate-level netlist into its nodes and edges, the PinGraph enables the model to capture latent patterns during EDA's place-and-route phase, thereby facilitating effective prediction of parasitic capacitance.

\minisection{Model Architecture}
The parasitic parameter prediction model consists primarily of an SMP-GNN, named ParaGate-Cap, and a readout layer. ParaGate-Cap's aggregator design draws inspiration from graph attention networks (GAT) \cite{velivckovic2017graph}, specifically designed to learn the underlying model of interconnect parasitic parameters from the local topological connectivity of the gate-level netlist. The aggregation process is governed by the following equations:
\begin{equation}
\small
\begin{split}
    \alpha_{ij} &= softmax_{j}((\vec{W}_q \vec{h}_i^{(k)})^\top \cdot (\vec{W}_{k} \vec{h}_j^{(k)}+\vec{W}_e \vec{e}_{ij})), \\
    \vec{m}_j &= \vec{W}_v \vec{h}_j^{(k)}+\vec{W}_e \vec{e}_{ij}  \label{eq:alpha_ij}, \\
    \vec{msg}_i &= aggr(\vec{h}_j^{(k)} \mid j \in \mathcal{N}(i)) = \sum_{j \in \mathcal{N}(i)} (\alpha_{ij} \cdot \vec{m}_j), 
\end{split}
\end{equation}
where, $\mathcal{N}(i)$ denotes the neighbors of node $i$, $\alpha_{ij}$ signifies the attention weight between node $i$ and node $j$, with $\vec{e}_{ij}$ representing the edge attribute between them. and $\vec{W}_q$, $\vec{W}_k$, $\vec{W}_v$, $\vec{W}_e$ are the weight matrices of the aggregator. The resulting aggregated message, $\vec{msg}_i$, is then fed into a gated recurrent unit (GRU)\cite{dey2017gate} to update the node's embedding.
\begin{equation}
    \small
    \vec{h}_i^{(k+1)} \leftarrow{ \text{GRU}(\vec{h}_i^{(k)}, aggr(\vec{h}_j^{(k)} \mid j \in \mathcal{N}(i)))}.
\end{equation}
These updated features are then passed through a multi-layer perceptron (MLP) \cite{popescu2009multilayer} readout layer for final regression prediction of the parasitic parameters.

\minisection{Pretraining and Fine-Tuning Framework}
Effectively transferring a parasitic-parameter prediction model across circuits of varying scales poses a significant challenge, as models trained solely on abundant small/medium-scale circuits fail to learn the unique topological structures of large-scale designs. To address this issue, we propose a tailored transfer-learning strategy that combines specialized sampling with fine-tuning.

As shown in \Cref{fig:framework}, we first pretrain on a dataset mainly composed of small and medium circuits, and subsequently fine-tune on large-scale circuits. In the pretraining phase, we construct features and labels for each circuit based on its netlist and standard parasitic exchange form (SPEF) data. Specifically, the features derive from the post-synthesis EDA report, while the labels capture post-place-and-route information. By training on a substantial volume of small/medium circuits, we optimize the graph regression pipeline—comprising aggregator, GRU, and MLP—yielding our pretrained ParaGate model.

During fine-tuning, we focus on large-scale circuits that contain extreme data points but have limited samples. To concentrate on high-discrepancy regions and avoid overfitting, we only sample and train on nodes with the most pronounced errors. We assume the primary adjustments for extreme values lie in the information-update mechanism and the readout MLP, while local neighbor aggregation patterns remain largely generalizable. Concretely, we decompose large circuits into sub-circuits, run inference with the pretrained model, and gather gradient information. We then select the top 20\% of sub-circuits with the largest average gradients for subsequent fine-tuning. In this stage, we only train the MLP and GRU parameters while keeping the aggregator fixed, thus preserving the pretrained model’s capacity to capture generalized local topological structures.




\subsection{Midstream - EDA Tool-Based Feature Enhancement}\label{3.2}

\minisection{PSPEF Construction and EDA Toolchain Integration}
To integrate ParaGate with the EDA toolchain, we constructed ParaGate's pseudo standard parasitic exchange format (PSPEF) file based on the parasitic parameter prediction model, strictly adhering to the SPEF standard format. This file primarily uses the D\_NET keyword to mark the predicted capacitance for each net while maintaining accurate connectivity.

\minisection{PSPEF-Based Feature Enhancement}
Utilizing the PSPEF file, the gate-level netlist is input into EDA tools to perform power analysis (PA) and static timing analysis (STA), thereby collecting enhanced features.

Leveraging PSPEF files and advanced EDA tool analysis reports, this step of ParaGate provides preliminary power and timing attributes for the gate-level netlist post-parasitic back-annotation. Crucially, it decouples the easily transferable parasitic parameters from the more challenging PPA attributes, furnishing the downstream task-specific calibration with physically meaningful and highly relevant enhanced features.

\subsection{Downstream - Global-information-based Calibration}\label{3.3}

At this stage, the downstream model learns from the enhanced features derived from the netlist topology. These features originate from EDA tool analysis reports generated after back-annotating predicted parasitic parameters, and their overall trends align with ground truth. Consequently, the model effectively captures the discrepancies between preliminary EDA tool reports and actual values, enabling more precise calibration of PPA predictions.

\minisection{Calibration Model Architecture}
The calibration for ATs relies on ParaGate-AT, while cell-wise power calibration uses ParaGate-Power. Similar to ParaGate-Cap, both ParaGate-AT and ParaGate-Power employ an SMP-GNN. These models each comprise an aggregator built with GAT, an updater constructed with GRU, and a final MLP readout layer to regress the calibration targets.
\begin{equation}
    \mathcal{L_{AT/POWER}} = \frac{1}{N_t}\sum{(\hat{y} - \frac{y_{AT/Power}}{y_{raw\_AT/raw\_Power}})^2} 
\end{equation}

\minisection{Calibration Model Features}
The calibration model's features are primarily composed of two sets. First, features reported by EDA tools from power and timing analysis on the netlist before parasitic back-annotation fully preserve the gate-level netlist's logical and electrical characteristics. Second, features from the EDA tool analysis reports are obtained after back-annotating the PSPEF file output by the parasitic parameter prediction model from the first stage. The latter set, by incorporating the impact of parasitic parameters, reflects the shifts in netlist characteristics like power and timing. Combining these two feature types allows the calibration model to more comprehensively understand netlist behavior, enabling precise correction of PPA prediction results.

%% file: doc/result.tex
\section{Experimental Results}
\label{sec:result}

We first detail our experimental setup in Section A, including training and test dataset configurations, model settings, and computational environment details. Subsequently, in Section B, we present the overall predictive performance of ParaGate along with comparisons to existing solutions. Finally, Sections C and D delve into ablation studies that analyze our staged prediction approach and the proposed fine-tuning framework.

\subsection{Experimental Setup}
This section details our experimental setup, covering dataset composition and generation, the code framework and model parameters, and the configurations for both comparative performance evaluations and ablation studies.

\begin{figure*}[tb!]
    \centering
    \subfloat[DFF's arrival time $R^2$ and MAPE comparison.]{ 
    \includegraphics[width=0.5\linewidth]{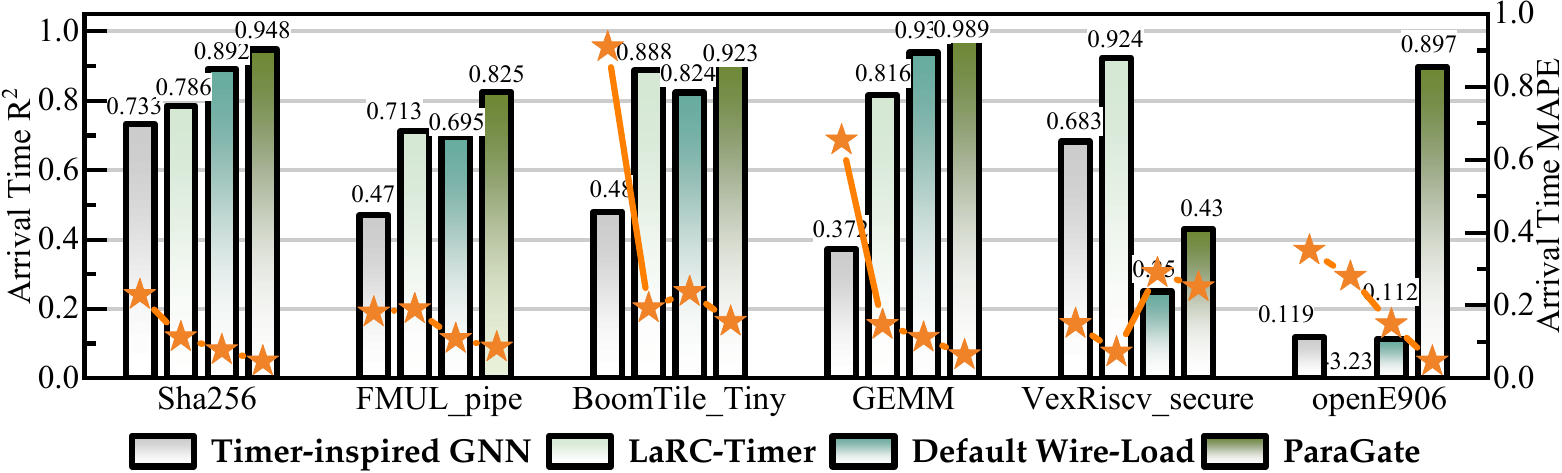} \label{fig:exp_1_at}}
    \subfloat[Cell-wise power $R^2$ and total power relative error comparison.]{ 
    \includegraphics[width=0.5\linewidth]{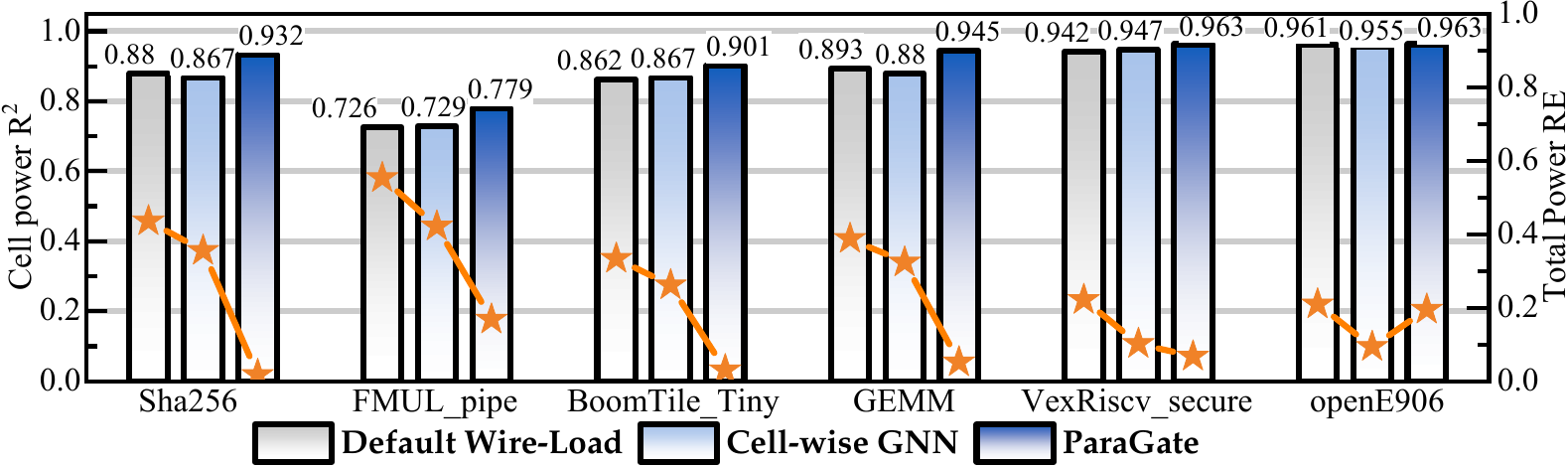} \label{fig:exp_1_power}}
    \caption{Overall Performance of ParaGate and Comparisons.}
    \label{fig:exp_1}
\end{figure*}

\minisection{Dataset Construction} The dataset, as detailed in \Cref{tab:circuit_info}, is partitioned into pretraining, training, and testing subsets. Specifically, the pretraining subset comprises 2004 small/medium-scale circuit instances, collected from public repositories such as GitHub and Hugging Face, with each instance ranging from 100 to 10,000 cells in size. The combined training and testing subsets consist of 14 open-source designs, with sizes ranging from 10,000 to 2,400,000 cells. To mitigate the impact of macro cells, units such as RAM and ROM were manually excluded from these designs.

\minisection{Circuit Data Processing using EDA Tools} Data processing utilized the TSMC 28nm technology library. The flow began with design synthesis via Synopsys Design Compiler's \textit{compile\_ultra} command. Place and route, performed using Cadence Innovus, then culminated in SPEF file generation. Subsequently, Synopsys PrimeTime generated final power and timing reports based on these SPEF files.


\minisection{Code Framework and model configuration} 
Leveraging PyTorch Geometric, we developed a parser for efficient netlist/feature-to-tensor conversion for GNNs, and GNN output to PSPEF for EDA tools. ParaGate integrates three GNNs (shared hyperparameters, differing in input feature dimensions). PinGraph node features are initialized to a 128-D latent representation. GNN layers aggregate information (128-D Q, K, V, E vectors; softmax-weighted sums) and update node latent vectors via a 128-D GRU. After 8 shared-weight GNN layers, latent vectors feed into a 3-layer MLP (256-D hidden) for final regression. Experiments were run on 8 × NVIDIA A100 GPUs.

\begin{table}[htbp]
\centering
\caption{Benchmark}
\label{tab:circuit_info}
\scriptsize  
\begin{tabular}{l|l|rrrr} 
\toprule
\multirow{2}{*}{Dataset} & \multirow{2}{*}{Circuit Name} & \multicolumn{4}{c}{Input Information} \\ 
& & \#node & \#edge & \#DFF & \\ 
\midrule
\multirow{6}{*}{test} 
& Sha256 & 19,346 & 72,180 & 1,112 & \\
& FMUL\_pipe & 60,815 & 236,749 & 2,395 & \\
& BoomTile\_Tiny & 157,022 & 515,210 & 19,726 & \\
& GEMM & 293,672 & 1,088,965 & 20,031 & \\
& VexRiscv\_secure & 420,674 & 1,497,966 & 74,650 & \\
& openE906 & 2,380,540 & 6,932,091 & 310,630 & \\
\midrule
\multirow{8}{*}{train} 
& VexRiscv\_small & 11,728 & 38,052 & 1,706 & \\
& RISCVmini & 20,865 & 74,506 & 1,624 & \\
& SodorCore & 24,983 & 94,250 & 1,668 & \\
& openE902 & 33,943 & 135,975 & 4,016 & \\
& FlexDPE & 67,998 & 278,129 & 2,113 & \\
& FPU & 135,725 & 573,539 & 1,543 & \\
& VexRiscv\_full & 502,154 & 1,509,160 & 76,232 & \\
& RocketTile\_Tiny & 973,877 & 2,870,165 & 168,555  & \\
\midrule
pretain & - & 2,340,911 & 7,424,599 & 210,631\\ 
\bottomrule
\end{tabular}
\end{table}




\subsection{\textbf{Overall Performance of ParaGate and Comparisons}}

\noindent\textbf{Comparison Settings.}
We train ParaGate on the training sets listed in \Cref{tab:circuit_info} and evaluate timing and power on the corresponding test sets. For benchmarking, we compare against the latest methods: reproduced versions of LaRC-Timer \cite{he2022accurate} and Timer-Inspired GNN \cite{guo2022timing} for timing predictions, and the Cell-wise GNN model from \cite{du2024powpredict} for power predictions, all with their physical parameter features removed. We also utilize an EDA tool with a wire-load model for timing and power predictions.


\minisection{Performance Comparison}
The comprehensive evaluation results are presented in \Cref{fig:exp_1}, where \Cref{fig:exp_1}(a) shows timing metrics and \Cref{fig:exp_1}(b) shows power metrics. Overall, ParaGate outperforms baseline models on most benchmarks in both AT and Power prediction except \textit{VexRiscv\_secure}, demonstrating strong generalization and modeling accuracy.
For AT prediction, ParaGate achieves higher $R^2$ and lower MAPE across nearly all designs, with a standout improvement on \textit{openE906}, where it raises $R^2$ from 0.119 to 0.897. 
For Power prediction, ParaGate improves cell-wise $R^2$ by 2\% to 5\% and reduces total power relative error to below 10\% for most designs. On \textit{BoomTile\_Tiny}, it achieves a relative error of only 0.909\%.
These results confirm ParaGate's superior performance and robustness across a wide range of design scenarios.
It is worth noting that in \Cref{fig:exp_1}, we mainly compare against the existing cross-stage methods. For a more comprehensive comparison, we also implemented a parasitic parameters predictor \cite{shen2024deep} and paired it with our subsequent two steps, and compared in \Cref{section:ablation1}.

\minisection{Inference-Time Analysis}
We also report the runtime for each of the three ParaGate inference time in \Cref{tab:sampling_performance}. For comparison, we adopt the method proposed in \cite{guo2022timing} for timing prediction and the approach in \cite{du2024powpredict} for power prediction as our baselines, measuring the prediction runtime and reporting the results in \Cref{tab:sampling_performance}.
On average, ParaGate is about 1.14× slower than the baseline. This overhead arises because the baseline employs single-step end-to-end inference, whereas ParaGate utilizes a three-step pipeline to boost prediction generalization. 
Thanks to employing a synchronous GNN in ParaGate—whereas existing end-to-end prediction models must resort to a computationally inefficient asynchronous GNN to emulate timing analysis—the relative runtime overhead compared to existing end-to-end prediction is acceptable.
For instance, on a 10k-cell circuit, the prediction time grows from 11.5s to 16.8s, and on the 2M-cell circuit \textit{openE906}, it grows from 864.9s to 881.3s. Notably, performing a full place-and-route flow to obtain performance metrics for this circuit can take tens of hours.

\begin{table}[htbp]
  \centering
  \scriptsize 
  \caption{Runtime (s) Comparison.} 
  \label{tab:sampling_performance} 
  \begin{tabular}{l|rrr}
    \toprule
    Circuit          & {Baseline} & {ParaGate} & {P\&R} \\ 
    \midrule
    Sha256           & 11.5        & 16.8 (7.9+0.9+8.0)  & 1067.0 \\ 
    FMUL\_pipe       & 67.9        & 85.6 (42.0+0.9+42.7)  & 4533.0 \\
    BoomTile\_Tiny   & 61.7        & 68.9 (34.0+2.4+32.5)  & 3362.0 \\
    GEMM             & 133.8       & 226.1 (104.0+4.0+118.1) & 7457.0 \\
    VexRiscv\_secure & 195.2       & 247.2 (122.0+6.8+118.4)  & 9501.0 \\
    openE906         & 864.9       & 881.3 (415.5+34.0+431.8)  & 47812.0 \\
    \bottomrule
  \end{tabular}
\end{table}

\subsection{\textbf{Ablation Study on Three-Step Framework}\label{section:ablation1}}
\begin{table}[htbp]
  \centering
  \scriptsize 
  \caption{Ablation Setting.} 
  \label{tab:ablation_Setting} 
  \begin{tabular}{l|ccc}
    \toprule
    Variants             & {Pretraining}  & {EDA-tool} & {Calibration} \\ 
    \midrule
    ParaGate w/o P & \textcolor{red}{\ding{55}} & \textcolor{darkgreen}{\checkmark} & \textcolor{darkgreen}{\checkmark} \\
    ParaGate w/o E & \textcolor{darkgreen}{\checkmark} & \textcolor{red}{\ding{55}} & \textcolor{darkgreen}{\checkmark} \\
    ParaGate w/o C & \textcolor{darkgreen}{\checkmark} & \textcolor{darkgreen}{\checkmark} & \textcolor{red}{\ding{55}} \\
    \bottomrule
  \end{tabular}
\end{table}
\minisection{Ablation Settings}
As shown in \Cref{tab:ablation_Setting},
``ParaGate w/o P'' adopts the capacitance prediction model from \cite{shen2024deep} and is trained directly on the training set as detailed in \Cref{tab:circuit_info}, without employing a pre-training and fine-tuning approach.
``ParaGate w/o E'' replaces the reasoning of EDA tools by training an AMP-GNN.
``ParaGate w/o C'' directly adopted the version of ParaGate without calibration.
\begin{figure}[htbp!]
    \centering
    \includegraphics[width=0.92\linewidth]{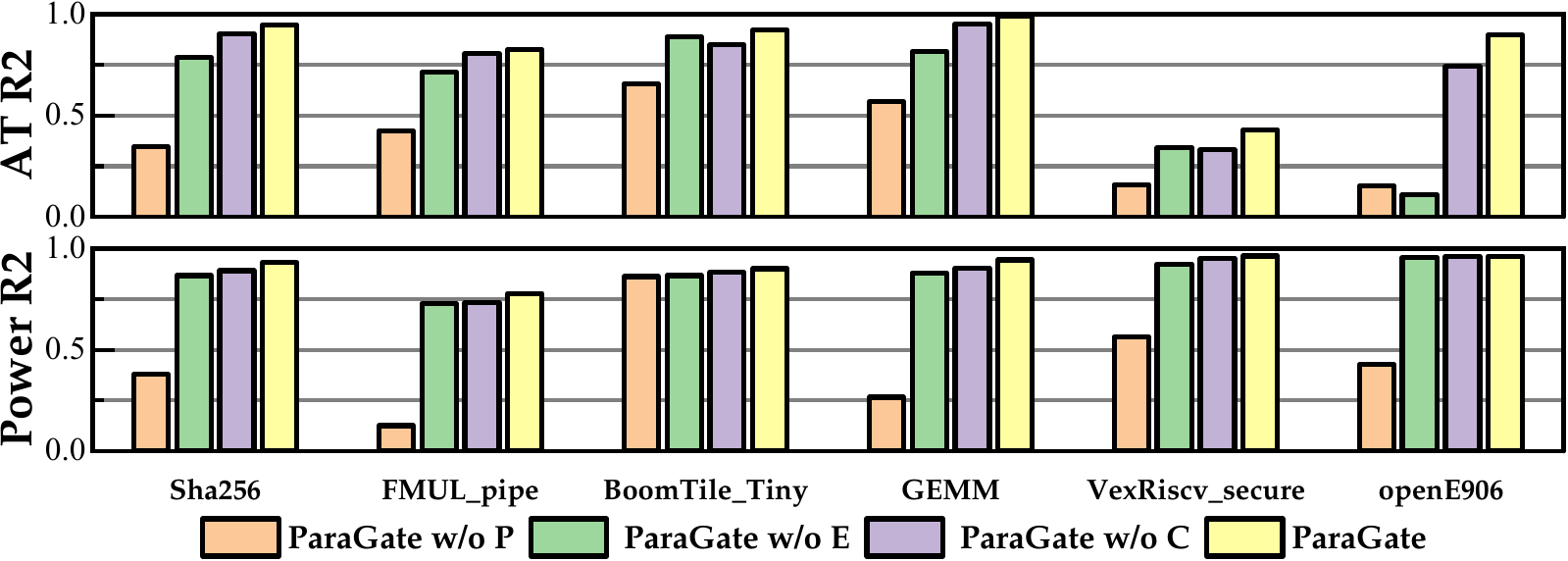} 
    \caption{ParaGate ablation study.}
    \label{fig:ablation}
\end{figure}

\minisection{Effect of Pretraining}
As shown in \Cref{fig:ablation}, ``ParaGate w/o P'' consistently underperforms notably due to overestimated capacitance, suggesting that training directly on large-scale circuits leads to overfitting and fails to capture realistic parasitic distributions. This highlights the necessity of pretraining on diverse small or medium scale designs.

\minisection{Effect of EDA-Tool Reasoning}
As depicted in  \Cref{fig:ablation}, ``ParaGate w/o E'' demonstrates lower prediction accuracy across all samples compared to the full version. 
The GNN predicting enhanced features within ``ParaGate w/o E'' inherently contains errors. This imprecision may arise from the conflict between neural networks' statistical fitting and the exactness of structured computation.

\minisection{Effect of Calibration}
``ParaGate w/o C'' shows lower accuracy than ParaGate. For example, on \textit{openE906}, calibration improves AT $R^2$ from 0.742 to 0.897, with typical gains ranging from 1\% to 7\%.

\subsection{\textbf{Ablation Study on Fine-tuning and Sampling Strategy }\label{section:ablation2}}

\minisection{Ablation Settings} We investigated the impact of sampling and fine-tuning on transfer learning performance using three methods based on our pre-trained parasitic parameter predictor: Grad-Freeze (20\% max-average gradient samples, frozen aggregator), Ran-Freeze (20\% random samples, frozen aggregator), and Grad-Update (20\% max-average gradient samples, full parameter update).

\begin{figure}[tb!]
    \centering
    \includegraphics[width=0.78\linewidth]{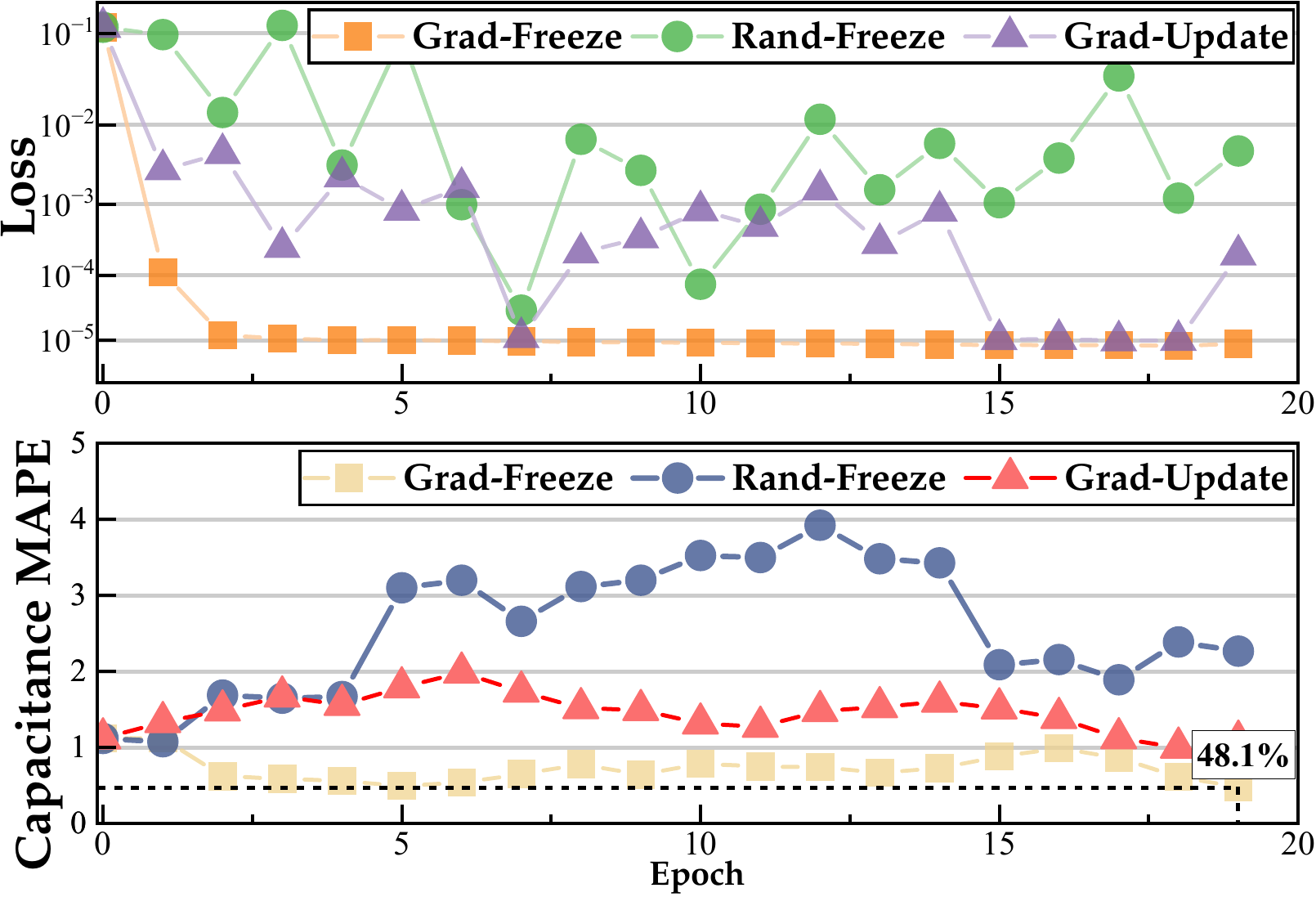} 
    \caption{Fine-tuning and sampling strategy ablation study.}
    \label{fig:ablation2}
\end{figure}

\minisection{Effect of Gradient-Based Sampling}
\Cref{fig:ablation2} illustrates that Rand-Freeze exhibits a more oscillatory loss curve and converges to a poorer accuracy compared to Grad-Freeze. This observation suggests that indiscriminately selecting training samples is ineffective in enhancing model capability. Conversely, gradient-based sampling proves beneficial in identifying truly informative samples from the training set that are crucial for improving the model's generalization ability.

\minisection{Effect of Freezing Aggregator Parameters}
\Cref{fig:ablation2} also demonstrates that Grad-Freeze achieves a faster and more stable convergence rate on the loss curve compared to Grad-Update. This indicates that maintaining the aggregator parameters fixed contributes to training stability. Furthermore, as shown in \Cref{tab:cap_result}, Grad-Freeze yields superior prediction accuracy during testing compared to Grad-Update, signifying that fixed aggregator parameters are conducive to enhancing the model's generalization capability.

\begin{table}[htbp]
  \centering
  \scriptsize 
  \caption{Capacitance MAPE.} 
  \label{tab:cap_result} 
  \begin{tabular}{l|rrr}
    \toprule
    Circuit     & {Grad-Freeze} & {Rand-Freeze} & {Grad-Update} \\ 
    \midrule
    Sha256           & \textbf{68.7}\% & 101.0\% & 93.5\% \\ 
    FMUL\_pipe       & \textbf{43.4}\% & 125.9\% & 123.5\% \\
    BoomTile\_Tiny   & \textbf{32.7}\% & 166.7\% & 142.9\% \\
    GEMM             & \textbf{108.8}\% & 166.7\% & 124.6\% \\
    VexRiscv\_secure & \textbf{29.9}\% & 176.5\% & 107.9\% \\
    openE906         & \textbf{39.1}\% & 237.8\% & 117.4\% \\
    \midrule
    Average          & \textbf{44.3}\% & 217.6\% & 116.7\% \\
    \bottomrule
  \end{tabular}
\end{table}

%% file: doc/conclu.tex
\section{Conclusion}
\label{sec:conclu}
This paper addressed the critical challenge of AI model generalization for early PPA estimation in unseen, large-scale circuits. We proposed ParaGate, a novel transfer learning framework that effectively decouples PPA prediction by combining scale-agnostic parasitic parameter learning with precise EDA tool-based reasoning. Our experiments demonstrate ParaGate's superior generalization performance with minimal fine-tuning, marking a significant step towards robust PPA assessment in complex circuit designs.